\documentclass{article}

\usepackage{arxiv}

\usepackage[utf8]{inputenc} 
\usepackage[T1]{fontenc}    
\usepackage{hyperref}       
\hypersetup{colorlinks=true,allcolors=blue}
\usepackage{url}            
\usepackage{booktabs}       
\usepackage{amsfonts}       
\usepackage{nicefrac}       
\usepackage{microtype}      
\usepackage{lipsum}
\usepackage{authblk}
\usepackage{graphicx}
\usepackage{subcaption} 

\title{Adversarial Patch Camouflage against Aerial Detection}

\author[*a]{\textbf{Ajaya Adhikari}}
\author[*b]{\textbf{Richard den Hollander}}
\author[a]{\textbf{Ioannis Tolios}}
\author[a]{\textbf{Michael van Bekkum}}
\author[c]{\textbf{Anneloes Bal}}
\author[c]{\textbf{Stijn Hendriks}}
\author[b]{\textbf{Maarten Kruithof}}
\author[d]{\textbf{Dennis Gross}}
\author[d]{\textbf{Nils Jansen}}
\author[e]{\textbf{Guillermo P\'{e}rez}}
\author[a]{\textbf{Kit Buurman}}
\author[a,f]{\textbf{Stephan Raaijmakers}}

\affil[a]{Dept. of Data Science, TNO, The Netherlands}  
\affil[b]{Dept. of Intelligent Imaging, TNO, The Netherlands}
\affil[c]{JIVC/KIXS/DataLab, Ministry of defense, The Netherlands}   
\affil[d]{Faculty of Science, Radboud University, The Netherlands} 
\affil[e]{Department of Computer Science, University of Antwerp, Belgium} 
\affil[f]{Centre for Linguistics, Leiden University, The Netherlands }

\begin{document} 
\maketitle
\let\svthefootnote\thefootnote
\let\thefootnote\relax\footnote{*Equal contribution\\ \phantom{.........}Corresponding authors: \texttt{ajaya.adhikari@tno.nl,\phantom{.} richard.denhollander@tno.nl}}
\let\thefootnote\svthefootnote
\addtocounter{footnote}{-1}\let\thefootnote\svthefootnote

\begin{abstract}
Detection of military assets on the ground can be performed by applying deep learning-based object detectors on drone surveillance footage.
The traditional way of hiding military assets from sight is camouflage, for example by using camouflage nets. However, large assets like planes or vessels are difficult to conceal by means of traditional camouflage nets. An alternative type of camouflage is the direct misleading of automatic object detectors. Recently, it has been observed that small adversarial changes applied to images of the object can produce erroneous output by deep learning-based detectors. In particular, adversarial attacks have  been successfully demonstrated to prohibit person detections in images, requiring a patch with a specific pattern held up in front of the person, thereby essentially camouflaging the person for the detector. Research into this type of patch attacks is still limited and several questions related to the optimal patch configuration remain open. 

This work makes two contributions. First, we apply patch-based adversarial attacks for the use case of unmanned aerial surveillance, where the patch is laid on top of large military assets, camouflaging them from automatic detectors running over the imagery. The patch can prevent automatic detection of the whole object while only covering a small part of it. Second, we perform several experiments with different patch configurations, varying their size, position, number and saliency. Our results show that adversarial patch attacks form a realistic alternative to traditional camouflage activities, and should therefore be considered in the automated analysis of aerial surveillance imagery. 
\phantom{----------------------------------------------------------------------------------------------------------------------------------------------------------------------------------------------------------------------------------------------------------------------------------------------------}
\end{abstract}


\section{INTRODUCTION}
\label{introduction}

\begin{figure}[!hbt]
\centering
\includegraphics[width=0.5\textwidth]{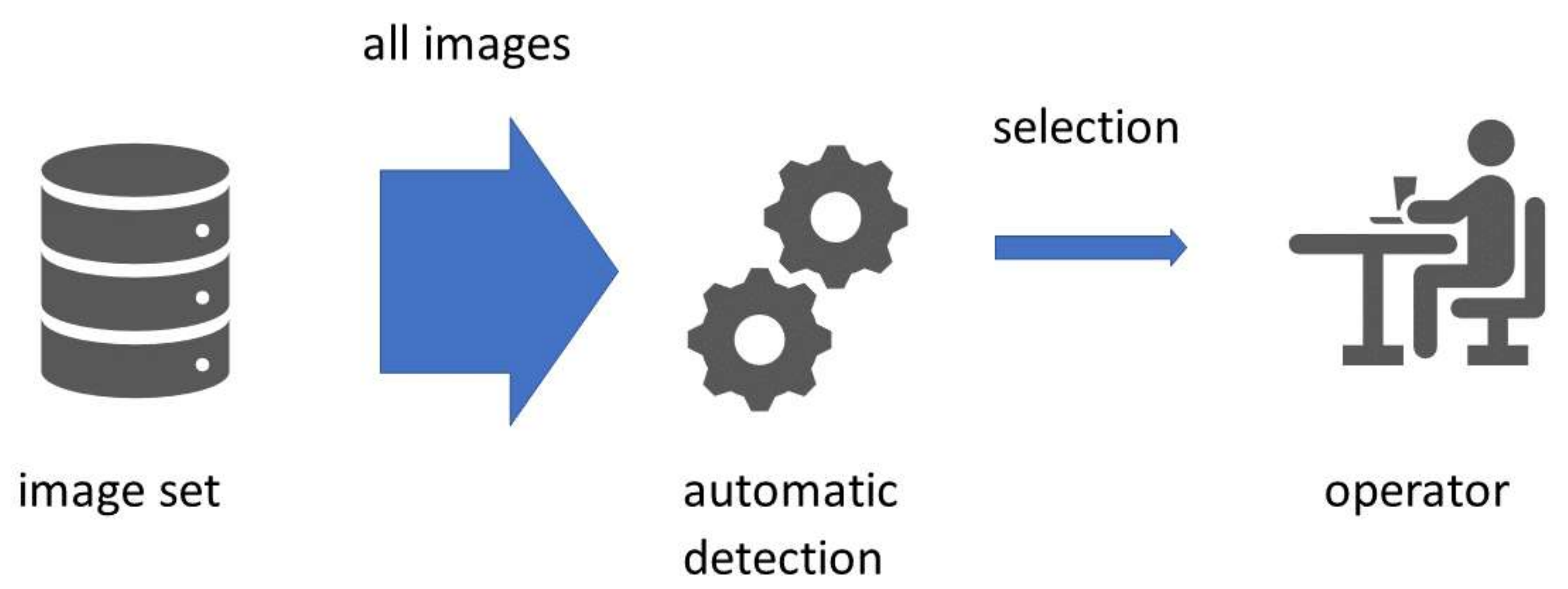} 
\vspace{0.5cm}
\caption{The potential of automatic detection algorithms in aerial data analysis.}
\label{introduction:fig1}
\end{figure}

The use of unmanned drones has increased considerably in recent years. Drones are used for reconnaissance and surveillance, and applied for spotting the presence of any activity on the ground. Due to the large quantity of image data that can be collected during drone surveillance, manual inspection of the image data is tedious. It is expected that image analysis in the near future will be  more and more automated~\cite{zhu2018visdrone,cvpr2019, iccv2017}. In particular, automatic object detection can be performed by deep neural networks trained on many examples of specific object classes, e.g. planes, vehicles, buildings etc. The result of this automatic analysis step is then presented to a human operator, who can confirm or reject the detection result (in case of false alarms), and use it as a cue for further analysis. The support of the analysis by machine learning methods can reduce the burden of manual inspection by several orders of magnitude (Fig.~\ref{introduction:fig1}). 

\begin{figure}[!hbt]
    \hspace{0.5cm}
    \begin{subfigure}{0.4\textwidth}
    \includegraphics[height=0.25\textheight]{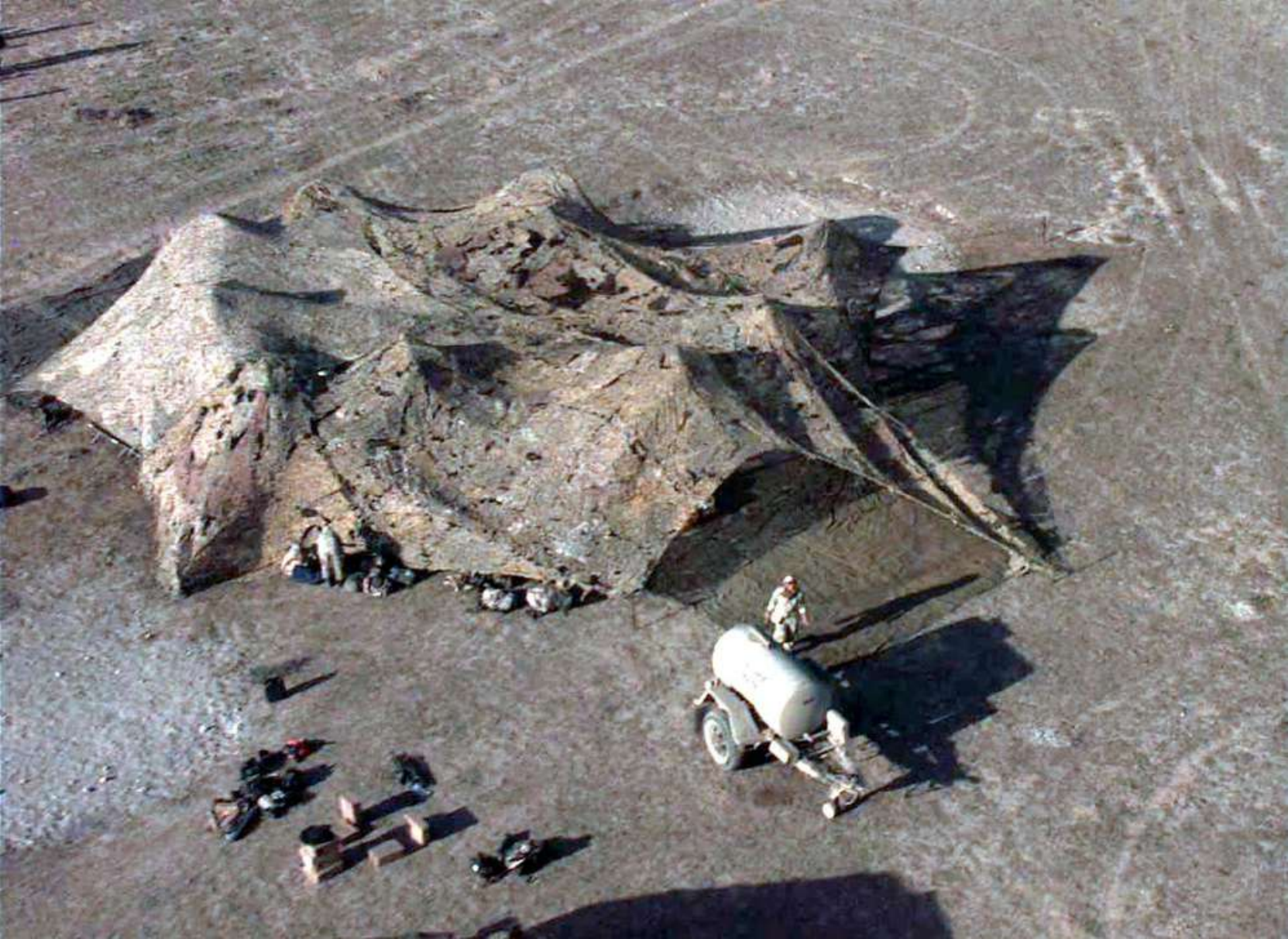} 
    \caption{Camouflage net}
        \label{fig:camouflage_net}
    \end{subfigure}
    \hfill
    \begin{subfigure}{0.4\textwidth}
        \includegraphics[height=0.25\textheight]{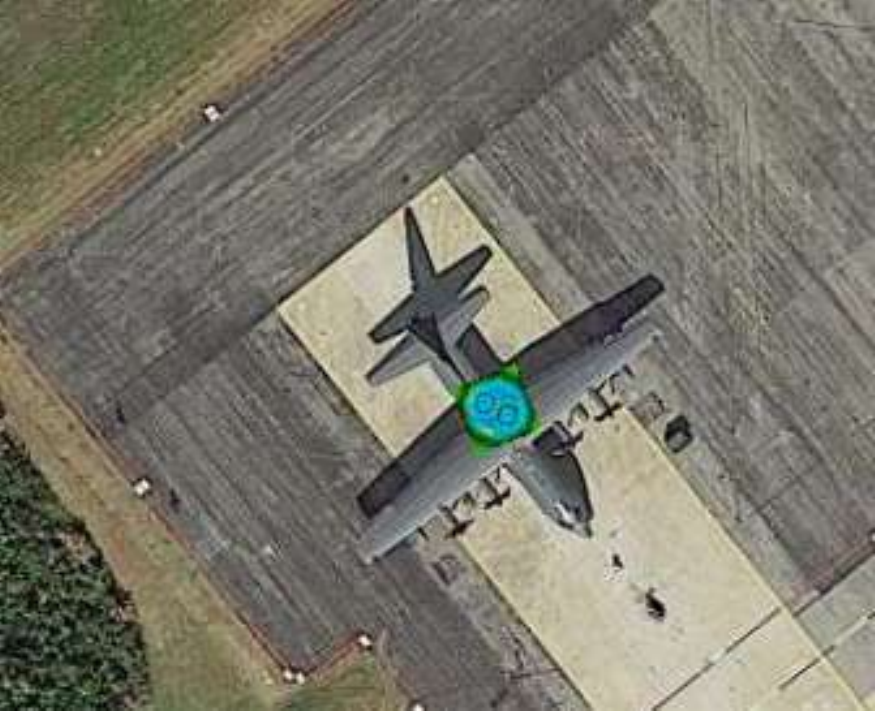}  
        \caption{Patch camouflage}
        \label{fig:patch_camouflage}
    \end{subfigure}
    \hspace{0.5cm}
    \caption{A camouflage net used for hiding an object\protect\footnotemark (left), and a plane with patch camouflage that can hide it from being automatically detected (right).} 
    \label{introduction:fig2}
\end{figure}

For military forces on the ground, the challenge is to conceal their presence as much as possible from discovery from the air. A common way of hiding military assets from sight is camouflage, for example by using camouflage nets, see Figure~\ref{fig:camouflage_net}. However, large assets like planes or buildings are difficult to conceal (quickly) by means of traditional camouflage. Moreover, large camouflaged objects may still show a recognizable outline, as it is a challenge to completely cover the object’s features.  

\footnotetext{source: \url{https://catalog.archives.gov/}}

In view of the process of data analysis as depicted in Fig.~\ref{introduction:fig1}, an alternative type of camouflage is the deception of automatic object detectors. Indeed, when the military asset is not detected by the automatic analysis, it will not be forwarded to the operator. Therefore, instead of hiding the object from manual inspection, the object can also be camouflaged by hiding it from automatic detection. This may not require full object coverage, but just a specific visual pattern --- laid out on or close to the object --- that affects the detectors’ decisions. In Figure~\ref{fig:patch_camouflage}, an example is shown of a plane overlaid with a patch. As will be shown in the remainder of the paper, this patch contains a specific learned pattern that can hide the plane from detection by the YOLO~\cite{redm2016} deep neural network.

The patch is an example of an adversarial attack on a neural network. In recent years, there has been a rise of neural networks for image classification and detection tasks, as their performance is unparalleled by alternative algorithms. At the same time, it was discovered that neural networks are sensitive to (small) perturbations of the input, raising questions regarding their reliability  in certain applications. In this paper, we propose to use a specific adversarial attack, namely the patch attack, as a means for camouflaging objects in aerial imagery against neural network-based detection. We further investigate how the effectiveness of the patch depends on parameters like size, position and color.

This paper is organized as follows. In Section~\ref{related_work} we give a short overview of related work on adversarial attacks for neural networks,  and discuss prior work on patch attacks. In Section~\ref{method} we describe our use of the patch attack for aerial camouflage, and we describe different possible patch configurations. In Section~\ref{experiments} we show the results of these patch configuration experiments on a large aerial image set. Finally, in Section~\ref{discussion} and \ref{conclusions} we discuss the results and present the conclusions and future work of this research.

\section{RELATED WORK}
\label{related_work}

Szegedy et al.~\cite{szeg2013} discovered that neural network classifiers are susceptible to specific distortions of the input images, causing the classifier output to be changed from the correct label into some arbitrary erroneous class label. Since this discovery, many researchers have worked on this type of adversarial attack and possible defenses against it, see Xu et al.~\cite{xu2020} for an overview. Most work on  adversarial attacks has been directed to adding distortions (noise) to images that are imperceptible for the human eye. More recently, research has also focused on real-world attacks, that is, changing the object or scene being photographed instead of modifying the pixel values in the photo. An example is the use of a small patch that is put on top of an object, which can be physically applied by means of a paper or poster.

The first adversarial attack based on an adversarial patch was introduced by Brown et al.~\cite{brow2017}. They developed a conspicuous patch that could be used for attacking image classifiers, without the need to do any image specific adjustments. The patch effectively altered the class label of the image, and could also be used as real-world attack. Liu et al.~\cite{liu2018} developed the so called DPatch attack specifically for object detectors. They showed that object detections from YOLO and Faster-RCNN can be effectively disabled by this patch attack. However, the patch constituted  the digital manipulation of a fixed group of pixels in the image corner, which does not correspond to a real-world attack. In addition, the manipulated pixel values could be outside the valid image range. Lee and Kolter~\cite{lee2019} improved upon DPatch by enforcing the pixel values to remain in the valid range and developing a real-world patch attack. Similar to DPatch, the patch did not overlap with the objects in the scene, but its influence was strong enough to block detections elsewhere in the image. The patch was developed and evaluated for YOLOv3. 

Adversarial patches have been targeted at specific applications. For example, Thys et al.~\cite{thys2019} have  proposed a real-world patch attack for the YOLOv2 detector that disabled person detection when a patch was held in front of a person. It was suggested that a printed patch on a t-shirt could be an effective way of hiding people from detection by surveillance cameras. This suggestion was picked up by Wu et al.~\cite{wu2019} who actually printed a pattern on a sweater that disabled person detections in YOLOv2/v3 and Faster-RCNN. Another application domain was the attack of face recognition systems with specific eyeglass patches by Sharif et al.~\cite{shar2016}, where the attack was effective in impersonating other individuals. An application in the military domain was the use of adversarial patches for camouflaging naval vessels by Aurdal et al.~\cite{aurd2019}. They created patches that could fool an image classifier in labelling a military vessel as type civilian. To the best of our knowledge, no adversarial attacks have yet been developed for camouflaging objects in aerial imagery, so as to prevent their detection.  

Along with research on adversarial patch attacks, research has also started on defense mechanisms against adversarial patches. Recent approaches include the method of Naseer et al.~\cite{nase2019} where patches for classification attacks are detected by local gradient magnitude and subsequently smoothed to reduce their effect, and Zhou et al.~\cite{zhou2020} where patches for detection attacks are found by an entropy measure and subsequently masked. A theoretical framework for provably effective defenses against patch attacks for classification was proposed by Chiang et al.~\cite{chia2020} and evaluated on MNIST and CIFAR-10 datasets. In the work of Levine et al.~\cite{levi2020} this approach was extended for use on the larger ImageNet dataset. However, these methods are designed for defenses against  classification attacks, while object detection in images will pose additional challenges. The work on defenses against adversarial patches for object detection is still limited, but will likely increase following the fast developments in the research of attacks.

\section{AERIAL CAMOUFLAGE}\label{method}
In this section we describe the patch training method and different  patch properties that make an attack practical for aerial camouflage. We focus on training and testing adversarial patches for hiding airplanes in aerial imagery. 

\begin{figure}[!hbt]
\centering
\includegraphics[width=0.7\textwidth]{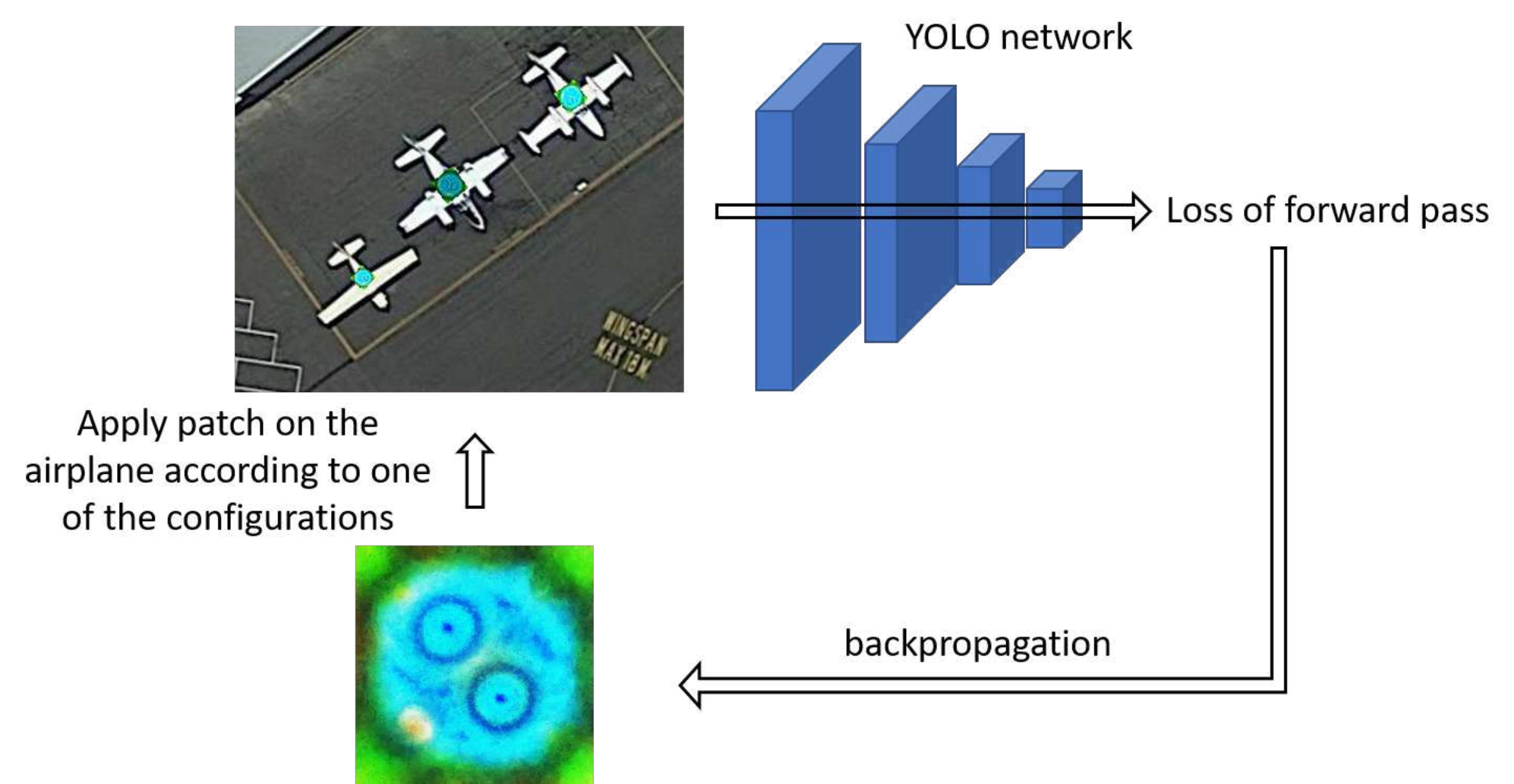} 
\vspace{0.5cm}
\caption{The adversarial patch training method~\cite{thys2019}. In each iteration, the current patch is applied to all planes in the batch of images, and in the forward pass the loss value is computed. The backpropagation pass is performed with fixed network weights, and the update is performed only to the patch pixels. The updated patch is subsequently used for the next image batch.}
\label{related:fig1}
\end{figure}

\begin{figure}[h]
    \begin{subfigure}{0.45\textwidth}
        \includegraphics[height=3cm]{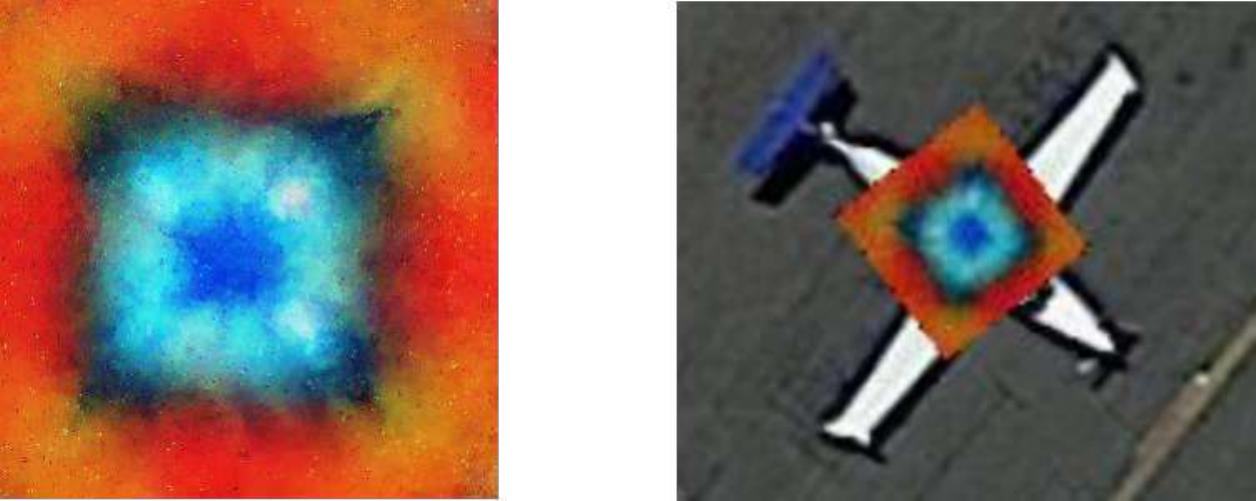} 
        \caption{Large patch}
        \label{fig:standard}
    \end{subfigure}
    \hspace{1.5cm}
    \begin{subfigure}{0.45\textwidth}
        \includegraphics[height=3cm]{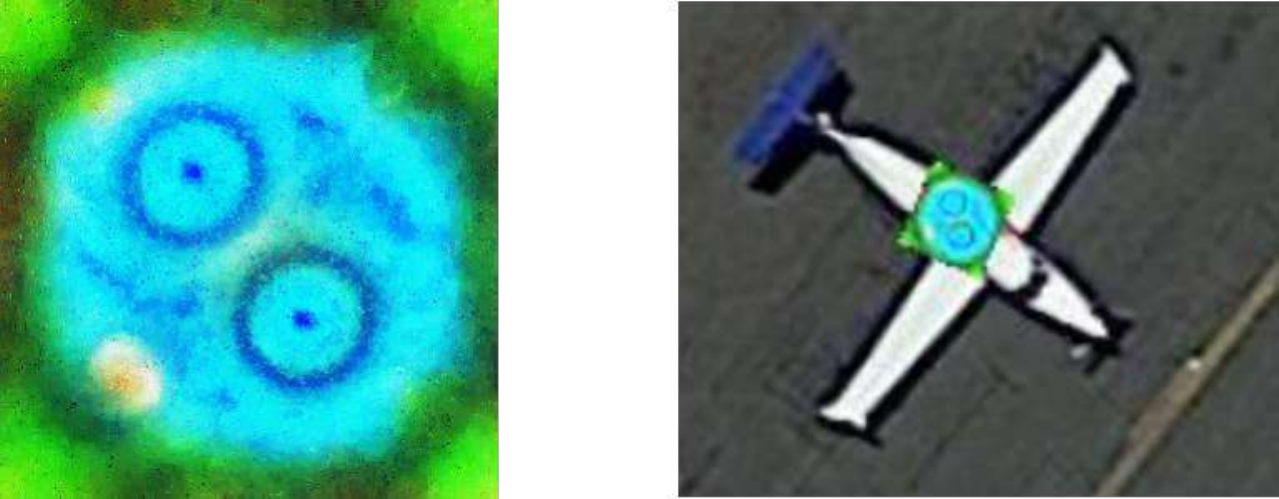}
        \caption{Small patch}
        \label{fig:small}
    \end{subfigure}
    \par\bigskip \bigskip
    \begin{subfigure}{0.45\textwidth}
        \includegraphics[height=3cm]{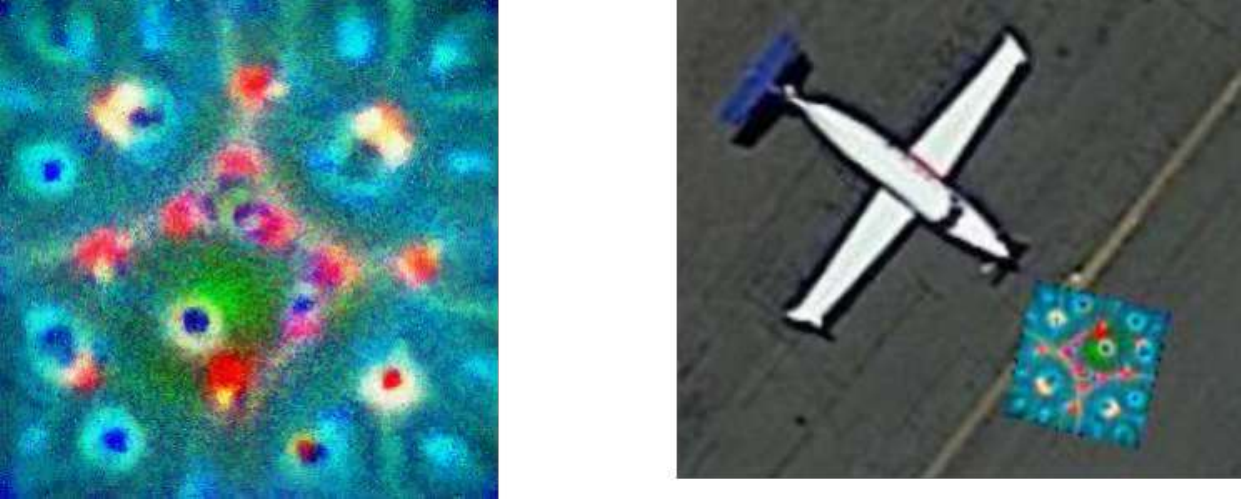}
        \caption{Large side patch}
        \label{fig:side}
    \end{subfigure}
    \hspace{1.5cm}
    \begin{subfigure}{0.45\textwidth}
        \includegraphics[height=3cm]{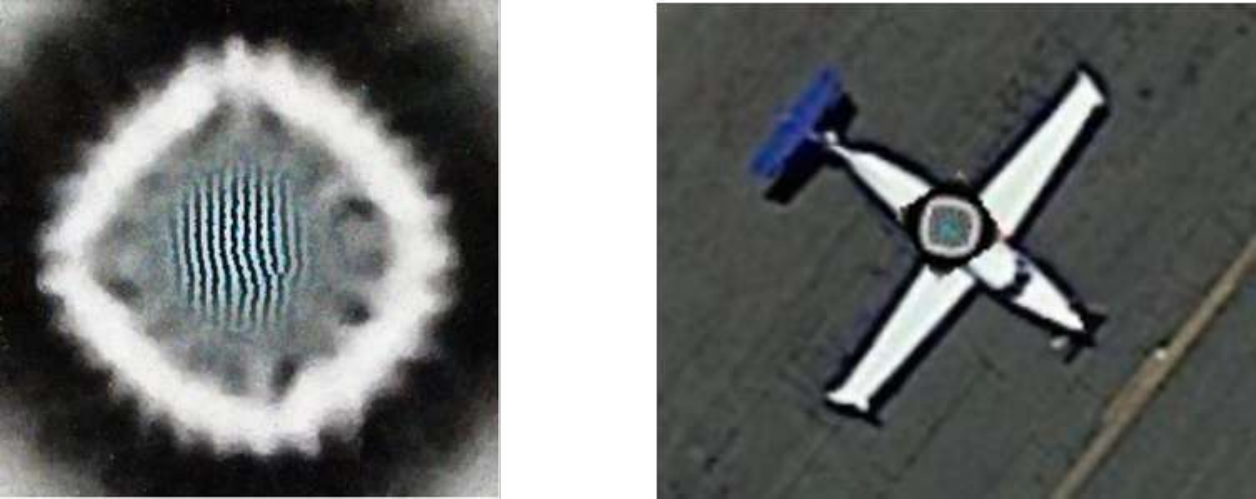}
        \caption{Small less colorful patch}
        \label{fig:less_saturated}
    \end{subfigure}
    \par\bigskip \bigskip
    \centering
    \begin{subfigure}{0.45\textwidth}
        \includegraphics[height=3cm]{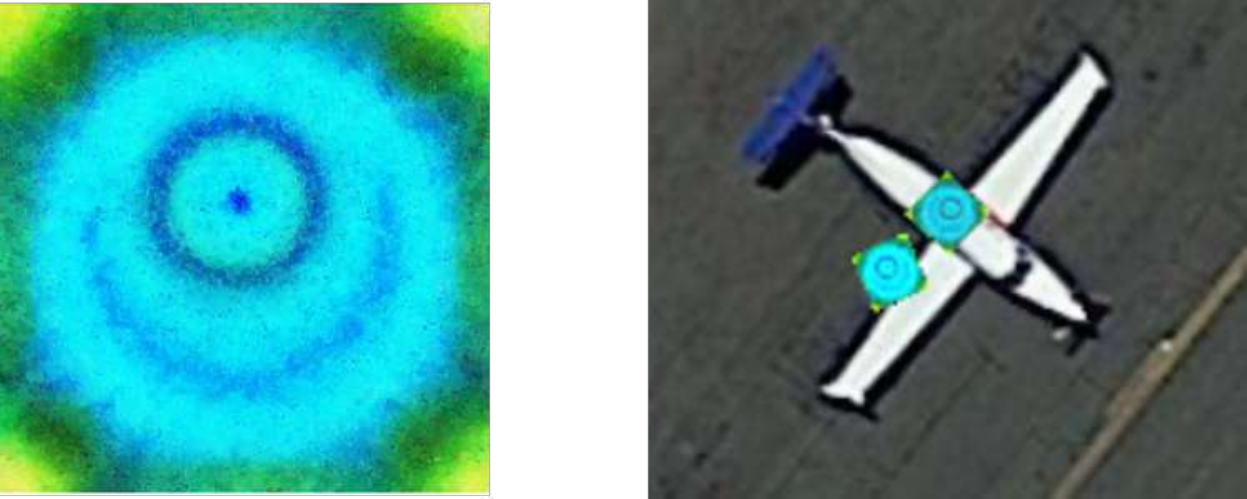}
        \caption{Two small patches}
        \label{fig:two_patches}
    \end{subfigure}
\caption{These figures illustrate the patch configurations (on the right of each figure) listed in table \ref{tab:patch_config} and corresponding final patch after training (on the left of each figure).}
\label{fig:patch_configurations}
\end{figure}

Our training method is built upon the work of Thys et al.~\cite{thys2019}, which is illustrated in Fig.~\ref{related:fig1}. 
In each iteration a batch of images containing airplanes is used for patch training. 
The current adversarial patch (initially starting with a random patch) is placed on the airplanes that have been annotated in the ground truth. 
The patches are scaled, rotated, corrupted with some noise and contrast stretched, such that they resemble real-life recording conditions, before placing them on the airplanes.
In our scenario, the patches are randomly rotated over 360 degrees since there is no fixed orientation in aerial images. 
The effect of this can be observed in the final trained patches (Figure~\ref{fig:patch_configurations}) which contain mostly circular symmetric patterns.
The YOLOv2 network is used for training, but has its weights fixed during back-propagation; only the patch is adjusted in each iteration. 
The following loss formula is used for optimization:

\begin{equation}
L = \alpha L_{nps} + \beta L_{tv} + L_{obj}
\label{eq:loss}
\end{equation}

where the non-printability score  $L_{nps}$ ensures printable colors in the patch and the total variation  $L_{tv}$ prevents the patch from becoming a noise pattern~\cite{thys2019}. The $L_{obj}$ term represents the maximum objectness score of an image in the YOLO output, and will therefore reduce the confidence of detections in the image.

\begin{table}[ht]
\caption{The configurations of the different experiments. The patch size  denotes the relative width and height of the patch in relation to the bounding box size of the airplane. } 
\label{tab:patch_config}
\begin{center}       
\begin{tabular}{|l|l|l|} 
\hline
\rule[-1ex]{0pt}{3.5ex}  \textbf{Experiment name} & \textbf{Patch size} & \textbf{Placement}  \\
\hline
\rule[-1ex]{0pt}{3.5ex}  {Large patch} & 0.2\,x\,0.2 & On top \\
\hline
\rule[-1ex]{0pt}{3.5ex}  {Small patch} & 0.1\,x\,0.1 & On top  \\
\hline
\rule[-1ex]{0pt}{3.5ex}  {Large side patch} & 0.2\,x\,0.2 & Next to  \\
\hline
\rule[-1ex]{0pt}{3.5ex}  {Small less saturated patch} & 0.1\,x\,0.1 & On top   \\
\hline 
\rule[-1ex]{0pt}{3.5ex} {Two small patches} & 0.075\,x\,0.075 (2x) & On top (2x)  \\
\hline
\end{tabular}
\end{center}
\end{table}

To make the attack practical, we considered several relevant properties of the patch namely its size, position, saliency and the number of patches.
The placing of the patch on the airplanes was varied during training, according to these patch properties, resulting in five different patch configurations; they are specified in Table~\ref{tab:patch_config} and illustrated in Figure~\ref{fig:patch_configurations}.
The rationale for the configurations in Table~\ref{tab:patch_config} is as follows.
First, the relative size of a patch w.r.t. the object is important for making the attack practical.
For example, if a patch has to cover a large part of an airplane to be effective, it has no added value compared to a camouflage net.
We therefore consider multiple patch sizes relative to the size of the plane (Figures~\ref{fig:standard} and~\ref{fig:small}).
Second, an attack can be more practical if a patch does not have to be placed on top of the object but next to it (Figure \ref{fig:side}).
Third, positioning multiple small patches (Figure \ref{fig:two_patches}) instead of one larger patch can be more practical, especially if the object has an irregular surface. The combined area of the two patches is comparable to the area of the single small patch.
Finally, it is useful if the patch is not salient, so that it is harder to detect by an automatic defense mechanism or more difficult to spot by the human eye.
To achieve this we steer the optimization process towards a less colorful patch (Figure \ref{fig:less_saturated}), using a colorfulness metric from Hasler and Suesstrunk~\cite{hasler2003measuring}, that was developed based on a user study in which participants rated images on their colorfulness.
This colorfulness metric is added as a saliency loss term $L_{sal}$ to the total loss value in Equation~\ref{eq:loss}:

\begin{equation}
L_{sal} = \sqrt{\sigma_{rg}^2 + \sigma_{yb}^2} + 0.3 * \sqrt{\mu_{rg}^2 + \mu_{yb}^2} 
\label{eq:saliency}
\end{equation}

\[
rg = R - G, \quad
yb = 0.5*(R+G)-B
\]
where $\mu$ and $\sigma$ are the average and standard deviation of the auxiliary variables, and $R, G$ and $B$ denote the values of the color channels of the patch.

\section{EXPERIMENTS}
\label{experiments}

This section describes the quantitative evaluation of the different patch configurations from Figure~\ref{fig:patch_configurations}. 

The experiments were performed on the DOTA dataset~\cite{Xia_2018_CVPR}, which contains about 2800 images  from Google Earth and  specific satellites. The dataset contains 16 annotated object categories like airplanes, ships and vehicles.
The large aerial images of varying sizes were divided into multiple smaller sub-images of 1024x1024 pixels, such that they could be fed to the object detection model in a consistent format. The collection of sub-images is subsequently split into two disjoint sets for training and testing, containing 2200 and 850 images with airplanes, respectively.
Adversarial patches were generated for the YOLOv2 object detector \footnote{\url{https://github.com/ringringyi/DOTA_YOLOv2}}, which was available pretrained on this dataset.

\begin{figure}[!hbt]
    \hspace{1cm}
    \begin{subfigure}{0.35\textwidth}
        \includegraphics[width=\textwidth]{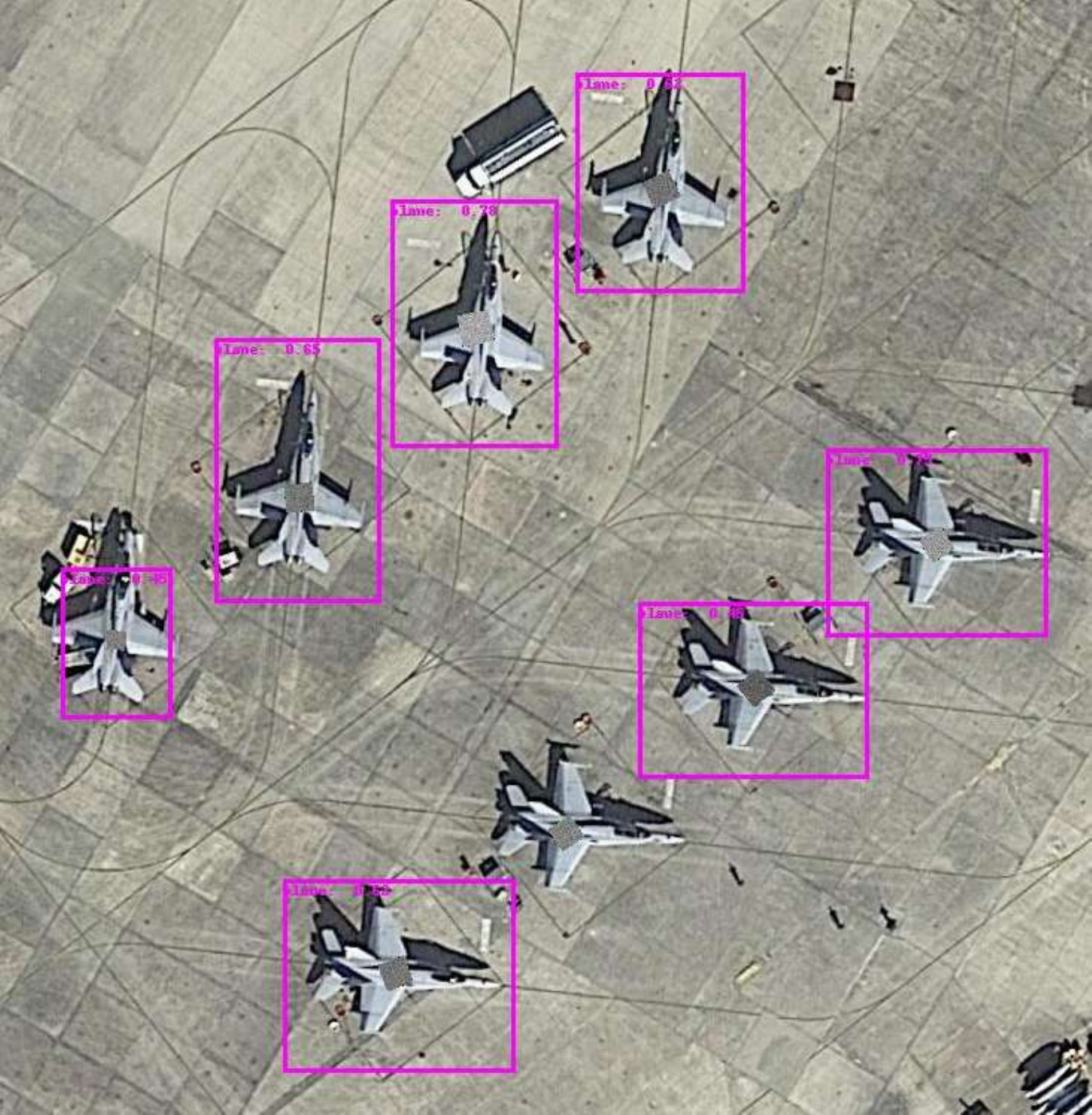} 
        \caption{Random patch}
        \label{fig:result_random}
    \end{subfigure}
    \hfill
    \begin{subfigure}{0.35\textwidth}
        \includegraphics[width=\textwidth]{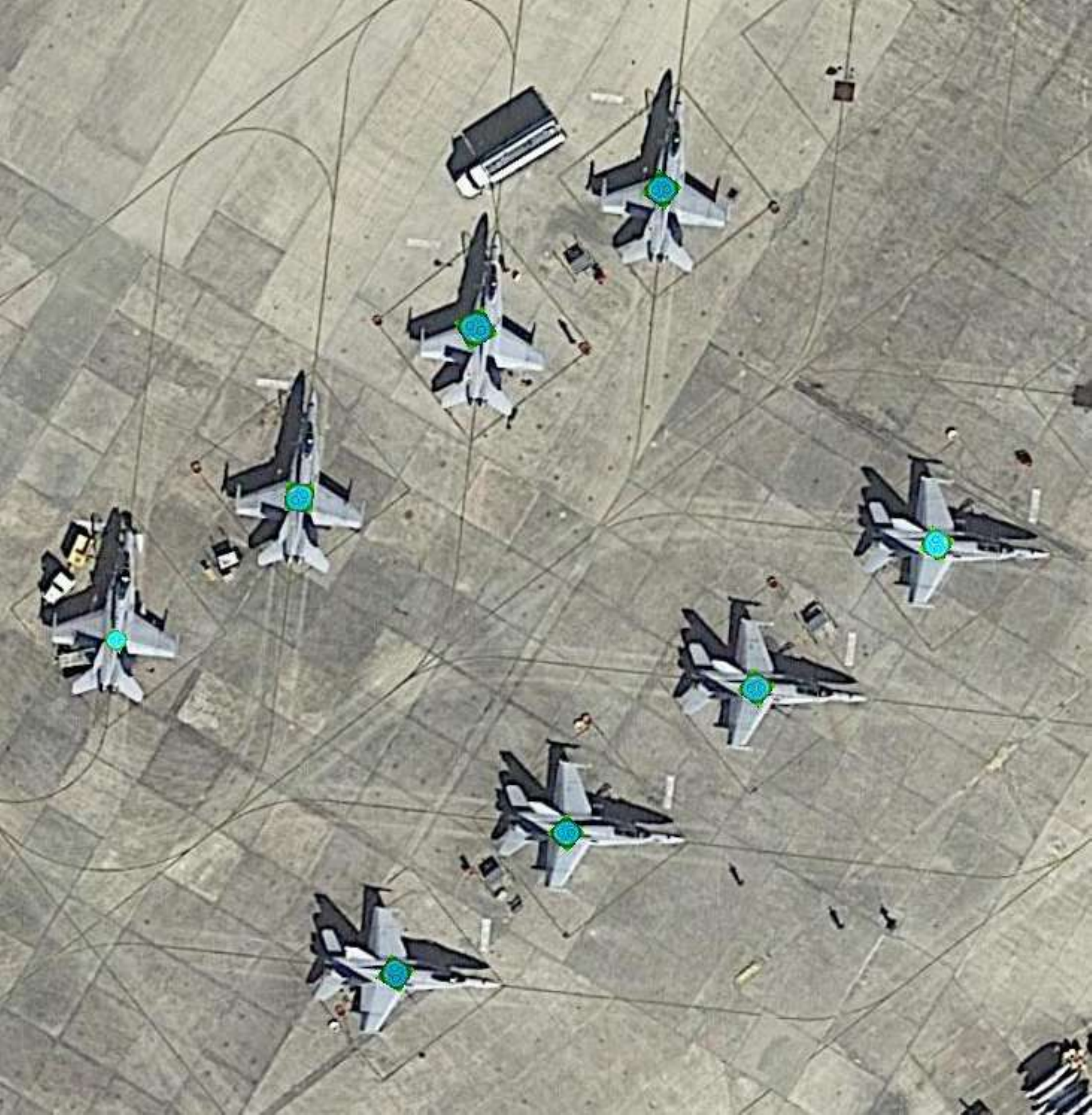} 
        \caption{Adversarial patch}
        \label{fig:result_patch}
    \end{subfigure}
    \hspace{1cm}
\caption{Images of jet fighters with detections overlaid for confidence threshold 0.4. The confidences of the detected jet fighters with random patches range from 0.45 to 0.78, with the undetected jet having a confidence of 0.23. The jets with adversarial patches are not detected at this threshold: confidences range from 0.01 to 0.04 with a single peak value at 0.14. }
\label{fig:random_patch_normal_patch}
\end{figure}

A patch was trained for 200 epochs on the train set for each configuration and evaluated on the test set in the same configuration (unless stated otherwise). The training of each patch took approximately 30 hours on an Nvidia GTX 1080Ti graphics card.
The detections found by the YOLO model on the raw images (without any patches) at confidence threshold 0.4 are taken as the ground-truth. Covering an object with any type of patch will naturally decrease its visibility, and
therefore each configuration is also evaluated with a random noise patch (similar to a camouflage net), which is taken as a baseline for comparison. These patches are generated using a uniform distribution over the RGB pixel range, resulting in patches with on average gray pixel colors.
Figure~\ref{fig:result_random} shows an example of an aerial image with such random noise patches and the corresponding detection result. 

\begin{figure}[!hbt]
    \begin{subfigure}{0.5\textwidth}
        \includegraphics[height=5cm]{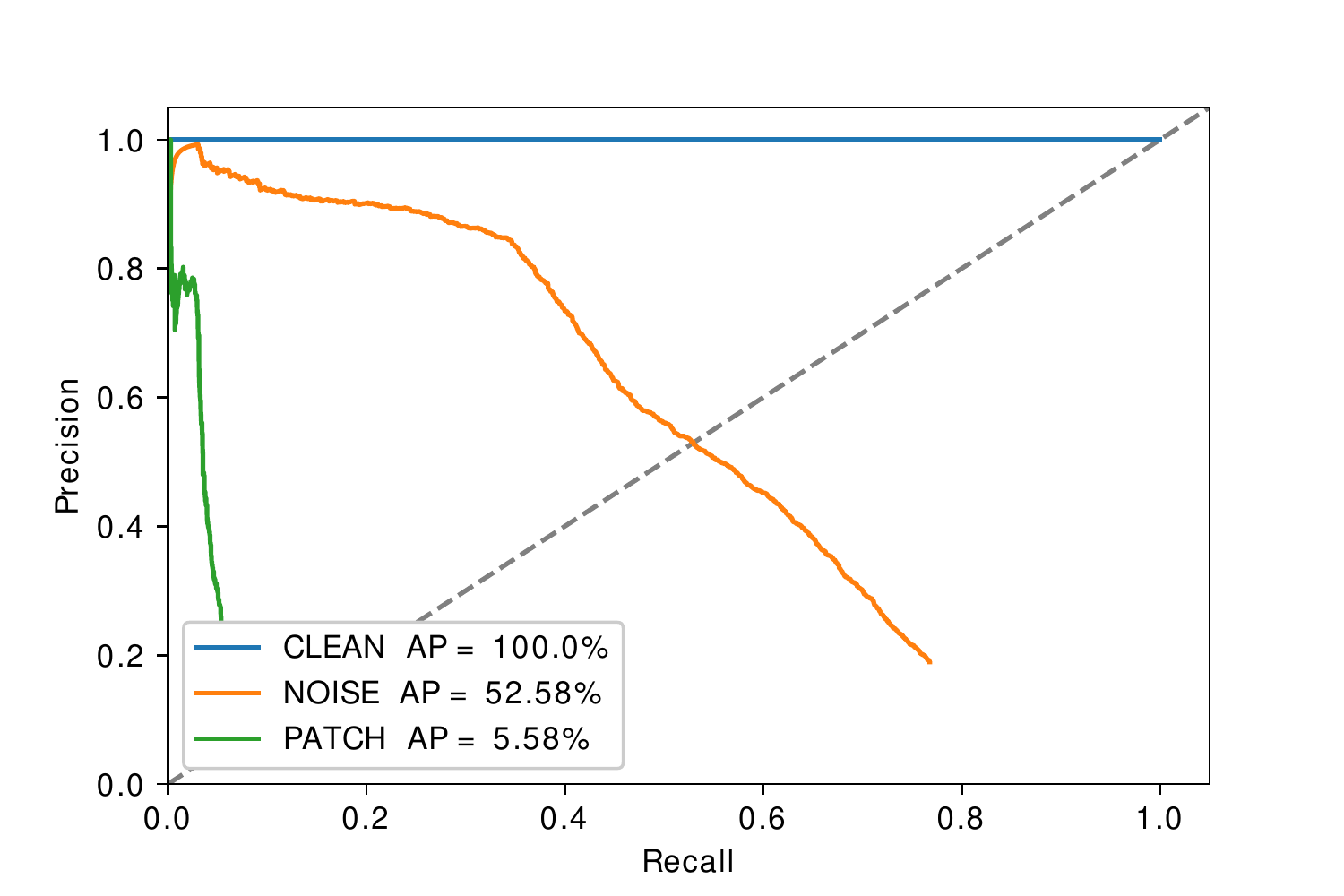} 
        \caption{Large patch}
        \label{fig:standard_evaluation}
    \end{subfigure}
    \begin{subfigure}{0.5\textwidth}
        \includegraphics[height=5cm]{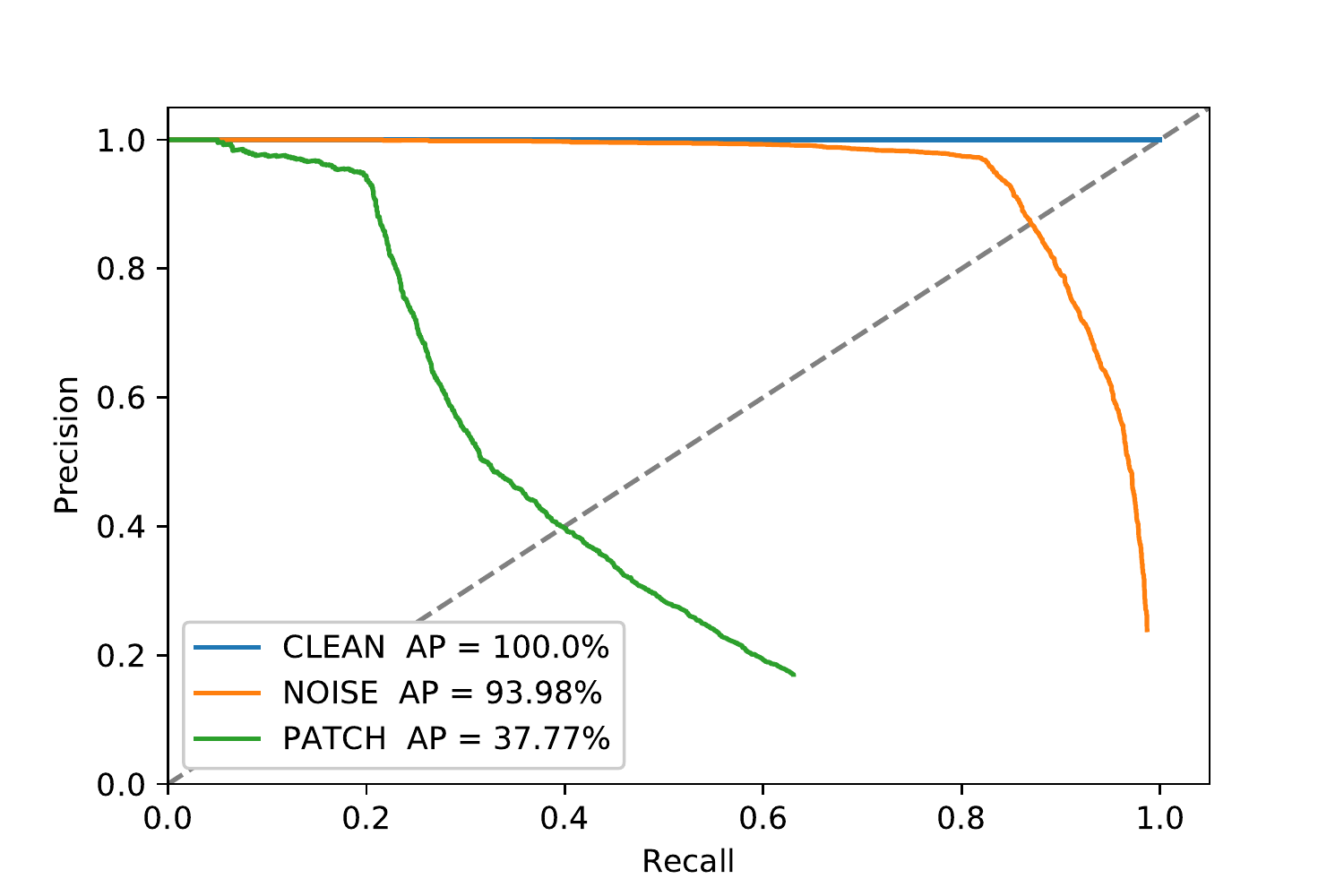}
        \caption{Small patch}
        \label{fig:small_evaluation}
    \end{subfigure}
    \par\bigskip 
    \begin{subfigure}{0.5\textwidth}
        \includegraphics[height=5cm]{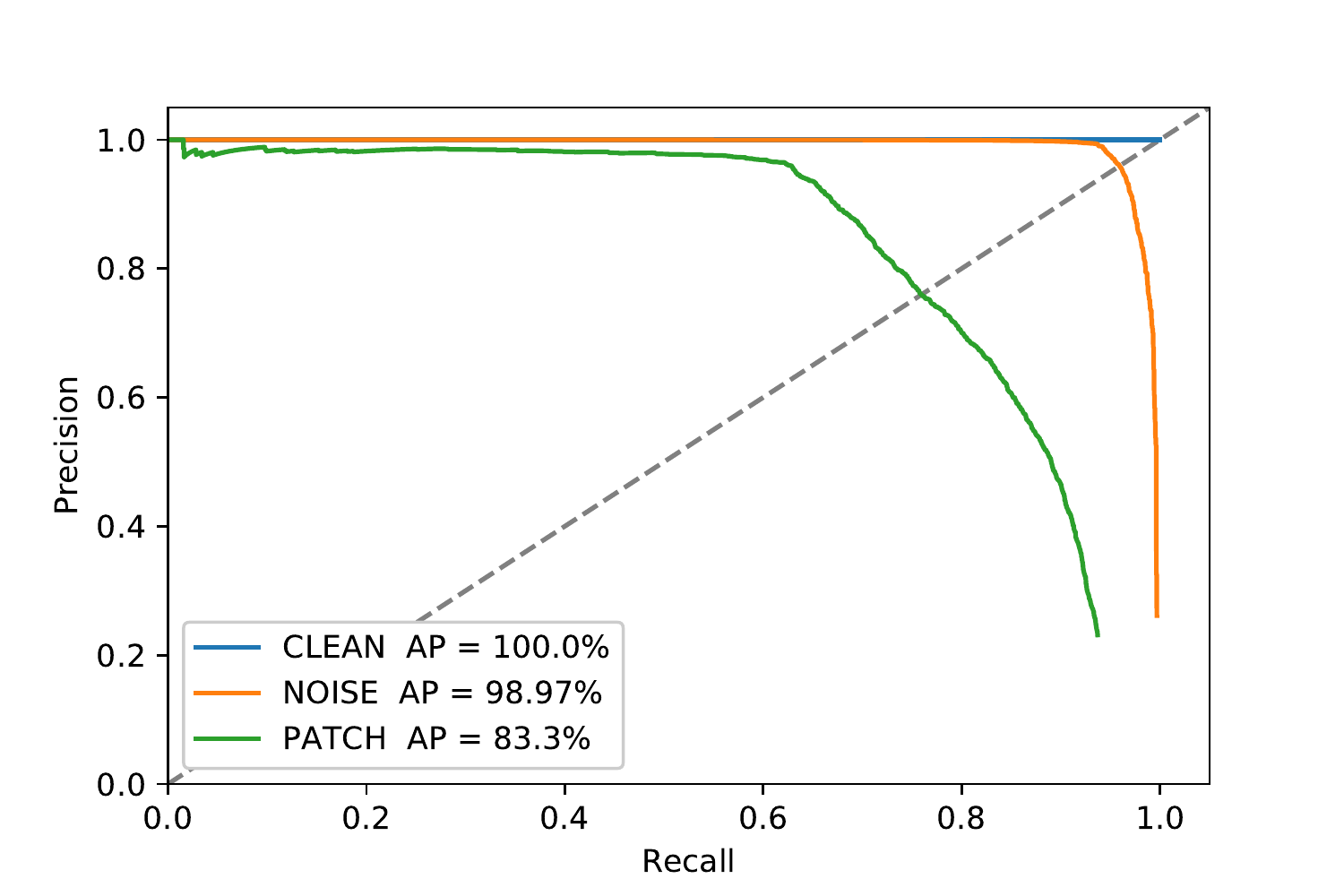}
        \caption{Large side patch}
        \label{fig:side_evaluation}
    \end{subfigure}
    \begin{subfigure}{0.5\textwidth}
        \includegraphics[height=5cm]{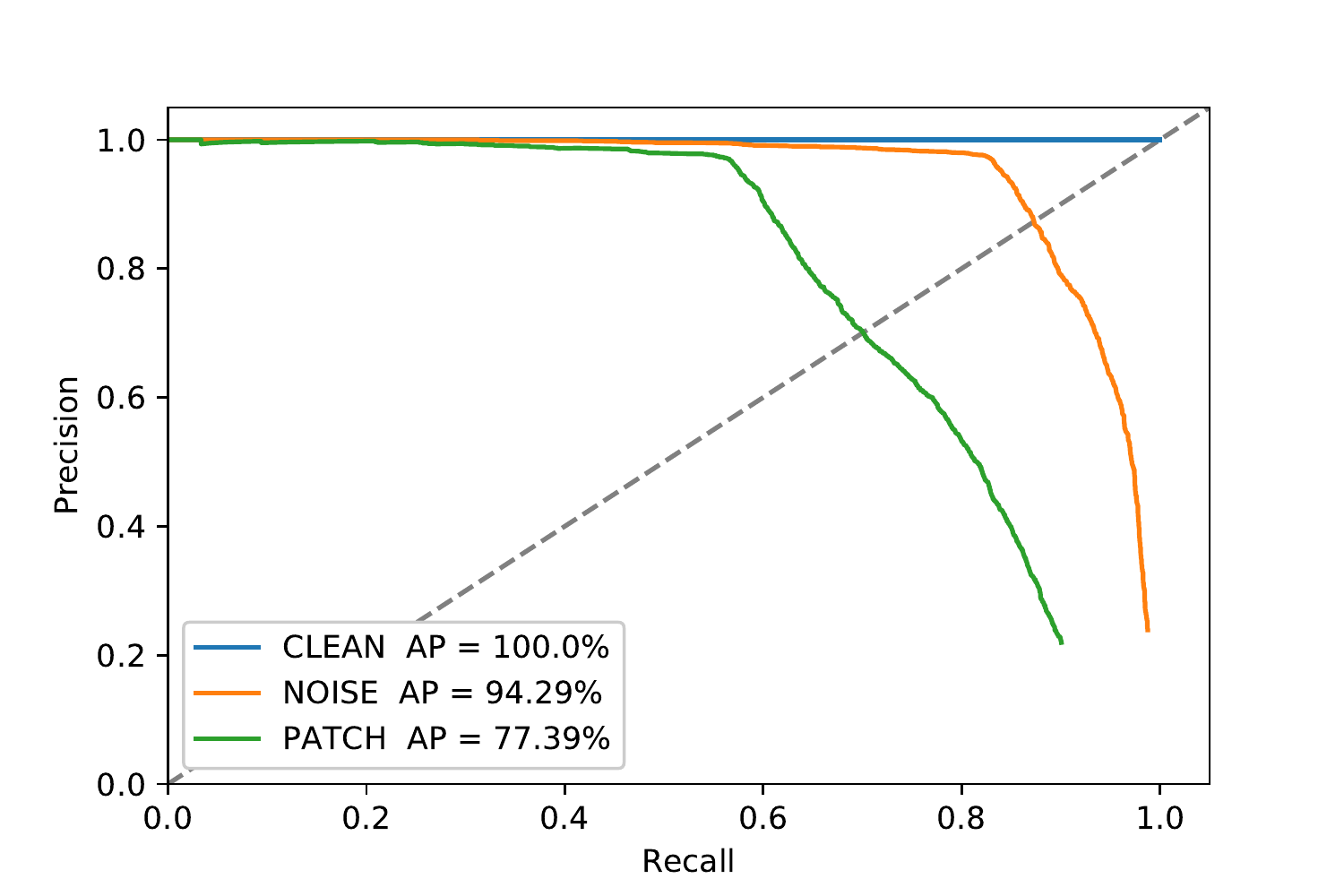}
        \caption{Small less colorful patch}
        \label{fig:less_saturated_evaluation}
    \end{subfigure}
    \par\bigskip
    \centering
    \begin{subfigure}{0.45\textwidth}
        \includegraphics[height=5cm]{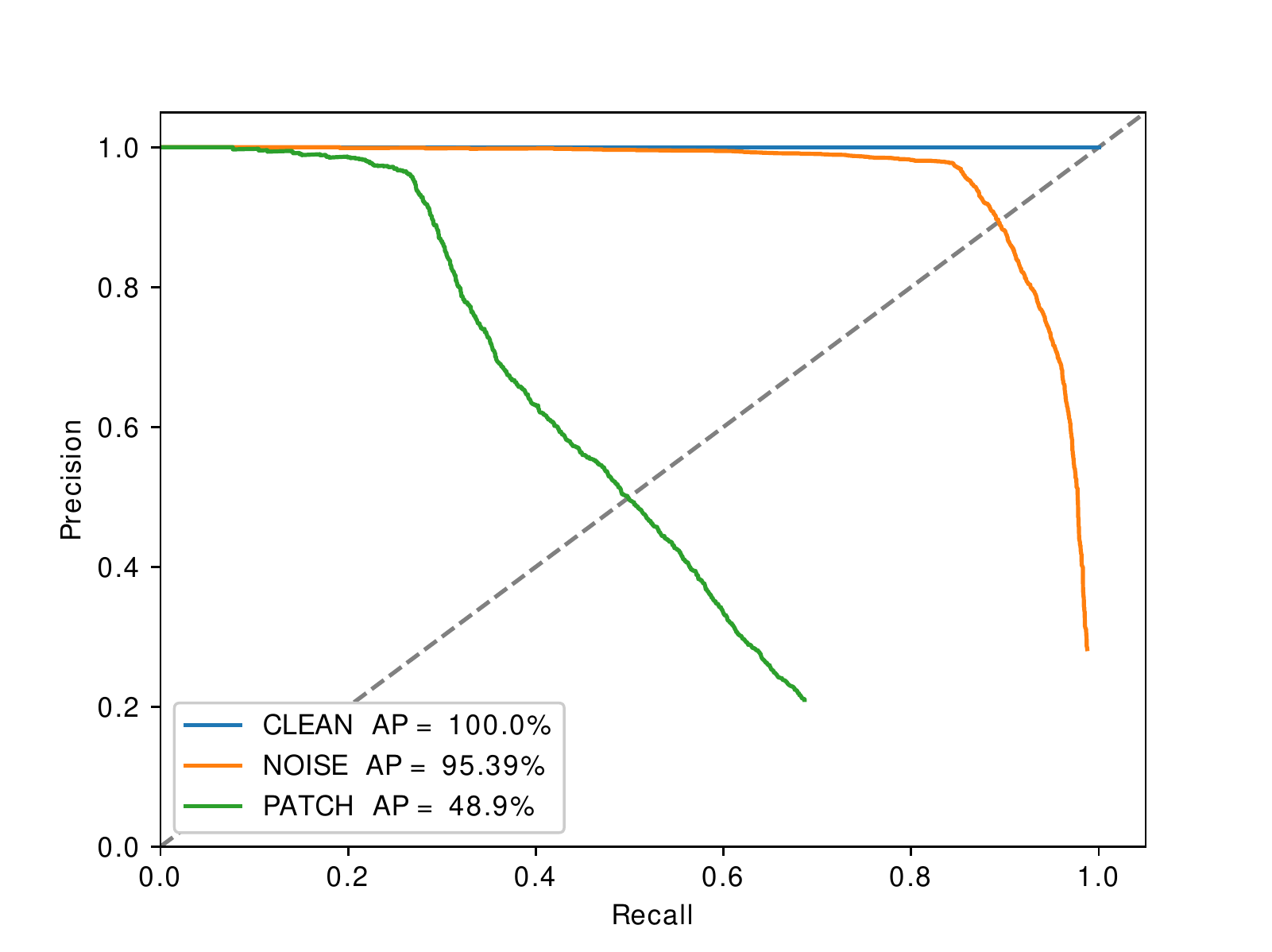}
        \caption{Two small patches}
        \label{fig:two_patches_evaluation}
    \end{subfigure}
\caption{The precision-recall curves for the different configurations listed in Table~\ref{tab:patch_config}. The CLEAN curve (blue) shows the evaluation of the detections on the raw images, but since these are taken as ground truth, it will by default equal 100\% precision. The NOISE curve (orange) is the evaluation on images with random patches on the planes, and the PATCH curve (green) for the images with the trained adversarial patches.}
\label{fig:patch_configurations_evaluation}
\end{figure}

The precision-recall curves of the different configurations are shown in Figure~\ref{fig:patch_configurations_evaluation}. Note that a patch attack is successful when the recall (detection rate)  is low for moderate or high values of the precision (in practice a high precision value is chosen to avoid too many false alarms). For example, when the detection threshold is set  such  that the precision is 80\% (on average 1 out of 5 detections is a false alarm), the large patch provides a recall value of 0\% so that all planes are effectively camouflaged. The large patch decreases the Average Precision (AP) of the model to 5.6\%, and is therefore very effective. However, the baseline noise patch also has a considerable effect (52.3\% AP) simply due to its size.
In contrast, the small patch achieves a bigger gain compared to the baseline noise patch: a decrease from 94.0\% AP to 37.8\% AP. The small patch pattern seems to be more powerful in that respect, see also Figure~\ref{fig:result_patch}.

Furthermore, we see that placing the patch next to the airplane is less effective than placing it on  top: the AP is lowered from 99.0\% to 83.3\% compared to the noise patch. A similar result is seen for the small less colorful patch, which has a much smaller AP decrease compared to the small patch. This suggests that the use of color plays an important role in fooling the object detection model.
The results further show that having two smaller patches does not lead to a more effective camouflage than having one small patch. 

\begin{figure}[!hbt]
    \begin{subfigure}{0.5\textwidth}
        \includegraphics[width=\textwidth]{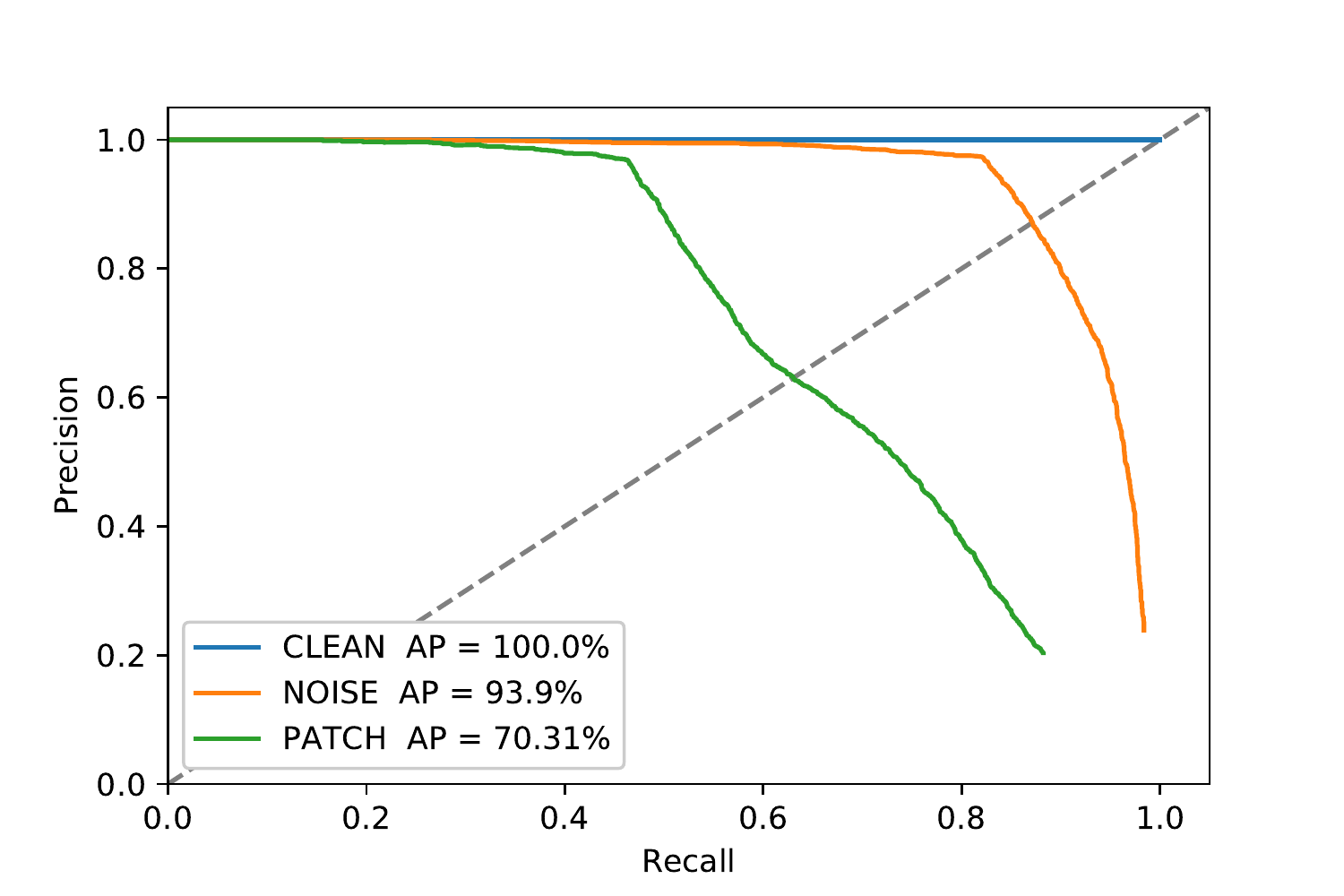} 
        \caption{The precision-recall curve}
        \label{fig:standard_as_smaller_evaluation}
    \end{subfigure}
    \hfill
    \begin{subfigure}{0.33\textwidth}
        \includegraphics[width=\textwidth]{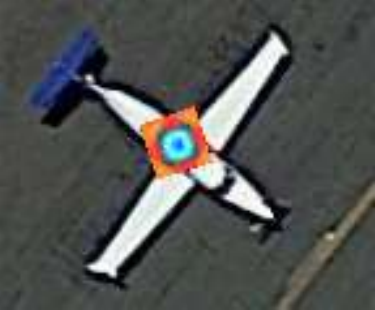} 
        \caption{Illustration of the placement of the patch during testing}
        \label{fig:standardpatch_as_smallpatch}
    \end{subfigure}
\caption{Evaluation and illustration of the patch that was trained with the large patch configuration, and placed on the airplane with the small patch configuration for testing.}
\label{fig:patch_configuration_standard_as_smaller_evaluatio}
\end{figure}

As additional experiment, we evaluated the effect of changing the patch configuration between training and testing phase. Figure~\ref{fig:patch_configuration_standard_as_smaller_evaluatio} shows the evaluation of the large patch that was tested in the small patch configuration.
The patch achieves a decrease of 23.6\% AP compared to the baseline noise patch.
However, the large patch tested in its original configuration has a larger decrease of 47\% AP. This suggests that the trained patches can still work in configurations that they were not trained for, but at the cost of a decrease in  effectiveness.

\section{DISCUSSION}
\label{discussion}

The experiments show that a relatively small patch can be used to camouflage an airplane against automatic detection. An advantage of this type of adversarial attack is that the patches are generic: only a single pattern is needed for camouflaging different types of planes. Still, an interesting question is whether it is beneficial to optimize the patch pattern for specific types of planes, like jet fighters versus airliners. In addition, there are more  patch configurations possible, like multiple patches next to the plane or multiple less-colorful patches, that may achieve better trade-offs between camouflage effect and practical usability.

We have not yet performed a  test with a printed patch on top of an actual airplane, in order to validate its performance in real-life. However, the patch training is set up in such a manner, that the resulting real-world patch is likely to have a similar effect. Experiments with  a printed adversarial patch for person camouflage (see e.g. Thys et al.~\cite{thys2019} and Wu et al.~\cite{wu2019}) have validated the real-world performance in a person detection scenario.

This work has been focused on the YOLO object detection network. There are more single-stage detection networks like YOLO, e.g. SSD~\cite{liu2016} and RetinaNet~\cite{lin2017}, and in addition two-stage networks like Faster-RCNN~\cite{ren2017} and Mask-RCNN~\cite{he2017}, which can all be used for aerial object detection. An adversarial patch should ideally be effective for all these network types, since it is unknown beforehand which object detector is used for analysing the drone footage. Wu et al.~\cite{wu2019} have found that  adversarial patches for person camouflage that are trained for YOLO are successful in fooling  Faster-RCNN  too. In addition, their ensemble training of an adversarial patch on multiple detectors simultaneously has shown generalization of the attack across networks. This strengthens the idea that the adversarial patch attack for aerial camouflage could be made effective for multiple detectors.

The detection of forces on the ground and the use of camouflage to prevent detection are in a continuous competition. In theory, new ways of real-world adversarial attacks on object detectors may spark the quest for appropriate defense mechanisms. Likewise, more robust detectors could initiate the search for better attacks. We therefore think that developments on both sides will go hand in hand. Next to the research on the attack, we are currently working on a framework that an operator can use for defense purposes against adversarial attacks.
This framework provides an automatic detection mechanism of potential adversarial attacks such that the operator can judge the correctness of such detections, and decide what follow-up action is needed for the analysis. 
This feedback is used to improve the detection of adversarial attacks and harden the object detector against similar attacks.
\section{CONCLUSIONS AND FUTURE WORK}
\label{conclusions}

In this paper we have applied adversarial patches for the purpose of camouflaging airplanes against neural network based detection in aerial imagery. A single patch pattern can be used for different types of planes. We have shown that relatively small patches, when compared to the size of the plane, can be used to camouflage the whole plane from automatic detection. From this point of view, the patches seem to be a more practical solution for camouflage than using camouflage nets. The results show that the patches do more than just occluding a small part of the plane: noise patches of similar size have a much smaller effect than adversarial patches.

There are several directions for future research. First, we think that further patch optimisation is possible for the investigated plane class or other object classes. Second, it should be investigated how good the patches can fool other neural networks than the ones they are trained on. An experiment with multiple object detection networks would therefore be interesting, including the concept of ensemble training of the adversarial patch. Third, evaluating the effectiveness of real-world patches on plane recordings is desired as a validation step of the digitally inserted patches.  Finally, the hardening of object detectors against potential patch attacks is an important research topic.

\section{ACKNOWLEDGEMENTS }
\label{Acknowledgements}
This research was supported by the NWO VWDATA P6 program "Verantwoorde Waardecreatie met Big Data" (VWData), project P6 ("Safe and Explainable AI"), funded by NWO (research number OND1364155).

\bibliography{report} 
\bibliographystyle{unsrt}  

\end{document}